\documentclass{article}

\PassOptionsToPackage{numbers, compress}{natbib}
\usepackage[preprint]{neurips_2023}

\usepackage[utf8]{inputenc} 
\usepackage[T1]{fontenc}    
\usepackage{hyperref}       
\usepackage{url}            
\usepackage{booktabs}       
\usepackage{amsfonts}       
\usepackage{nicefrac}       
\usepackage{microtype}      
\usepackage{xcolor}         
\usepackage[pdftex]{graphicx} 
\usepackage{floatrow}
\newfloatcommand{capbtabbox}{table}[][\FBwidth]
 \usepackage[font=footnotesize,skip=0pt]{caption}

\title{POP: Prompt Of Prompts for Continual Learning}

\author{
Zhiyuan Hu$^1$~~~
Jiancheng Lyu$^2$~~~
Dashan Gao$^2$~~~
Nuno Vasconcelos$^1$~~~
\smallskip 
\\
$^1$UC San Diego~~~
$^2$Qualcomm AI Research\footnote{Qualcomm AI Research is an initiative of Qualcomm Technologies, Inc.}~~~
\smallskip
\\
\{\tt\small{z8hu, nvasconcelos\}@ucsd.edu, ~ \{jianlyu, dgao\}@qti.qualcomm.com} 
}

\let\ACMmaketitle=\maketitle
\renewcommand{\maketitle}{\begingroup\let\footnote=\thanks \ACMmaketitle\endgroup}
\begin{document}

\maketitle

\begin{abstract}
Continual learning (CL) has attracted increasing attention in the recent past. It aims to mimic the human ability to learn new concepts without catastrophic forgetting. While existing CL methods accomplish this  to some extent, they are still prone to semantic drift of the learned feature space. Foundation models, which are endowed with a robust feature representation, learned from very large datasets, provide an interesting substrate for the solution of the CL problem. Recent work has also shown that they can be adapted to specific tasks by prompt tuning techniques that leave the generality of the representation mostly unscathed. An open question is, however, how to learn both prompts that are task specific and prompts that are global, i.e. capture cross-task information. In this work, we propose the Prompt Of Prompts (POP) model, which addresses this goal by progressively learning a group of task-specified prompts and a group of global prompts, denoted as POP, to integrate information from the former. We  show that a foundation model equipped with POP learning is able to outperform classic CL methods by a significant margin. Moreover, as prompt tuning only requires a small set of training samples, POP is able to perform CL in the few-shot setting, while still outperforming competing methods trained on the entire dataset.
\end{abstract}

\vspace{-10pt}
\section{Introduction}
\vspace{-10pt}


\textit{Continual learning } (CL), also called life-long learning, aims to replicate the human ability to learn new tasks without forgetting previously learned ones.  This, however, is not an easy goal for neural networks trained by gradient descent. Under gradient descent training, fine-tuning a model on new tasks tends to completely destroy the parameters and feature space learned for old tasks. Hence, when presented with new training data, neural networks easily forget what they have learned in the past, a problem known as \textit{catastrophic forgetting}. This has induced substantial research in CL, where two main classes of methods have been proposed to mitigate the problem. One approach is to constrain the parameters of the model learned on a new task to deviate little from their values before this learning, e.g. by using distillation~\cite{icarl, podnet, dark, lucir} or parameter regularization~\cite{ewc, nscl} methods.  A second approach is to fix the existing network and learn a new model for the new task, whose features are then concatenated to those of the latter. 


\begin{figure}\RawFloats
    \centering
    \includegraphics[width=0.8\linewidth]{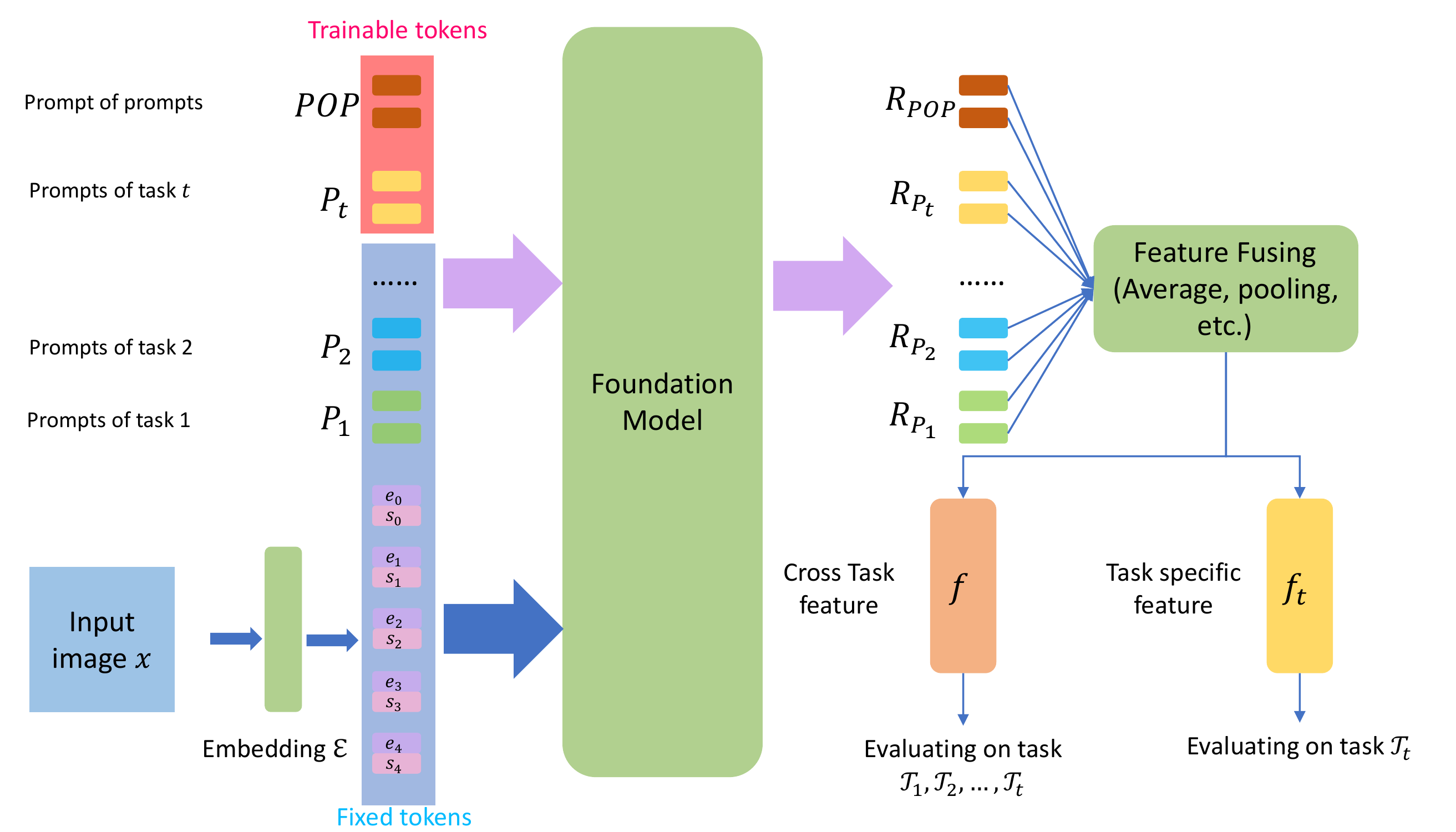}
    \caption{The POP model for CL based on a foundation model. A group of task-specific prompts $P_t$ is learned per task $t$. An extra group of global prompts, denoted as $POP$, is learned to integrate information across tasks. At step $t$ only $P_t$ and $POP$
    are updated. By fusing the transformer outputs corresponding to these prompts, the CL model learns both a task-specific representation $f_t$ and a global cross-task representation $f$, which  is suitable to solve the CL problem.}
    \label{fig:model}
\end{figure}

While these approaches have achieved satisfactory results, they are still prone to forgetting, have potentially unbounded memory and time complexity as the number of tasks grows, and exhibit a substantial gap with the respect to the joint learning of all tasks. All of these properties are unlike human CL, which is also very efficient, allowing humans to learn effectively in the few-shot regime, i.e. from few examples per class. This is not totally surprising, since humans leverage a very robust feature representation, produced by a brain optimized by a long evolutionary process. On the other hand, CL methods are usually limited to learn a model from scratch, and thus lack a representation that can easily generalize to new tasks. However, this constraint is artificial for many applications, where models pretrained on large datasets are readily available.


The issue becomes even more relevant in light of the recent introduction foundation models~\cite{foundation_survey, gpt, ldm, sam}. These are models trained on very large scale datasets, and thus endowed with a feature representation rich in semantic information for a very wide range of classes. Moreover, since most foundation models are transformers~\cite{transformer}, they can be efficiently adapted to specific tasks by prompt tuning techniques~\cite{prompt_tuning}, which only learn a small set of parameters. Hence, foundation models provide a foundation of CL that is closer to that of human learning: a generalizable feature representation with great potential for few-shot learning.

In this work, we test this hypothesis, by investigating the performance of prompt-based foundational model adaptation as a solution to CL. We consider the class incremental learning problem, where tasks consist of different classes of a classification problem. We propose a new continual prompting technique, denoted as the {\it Prompt of Prompts\/} (POP) model, illustrated in Figure~\ref{fig:model}. This consists of augmenting the set of tokens extracted from the image to classify with two sets of prompts. The first are task prompts $P_t$, which are learned per task $t$ and then frozen, and encourage the foundation model to discriminate {\it locally,\/} between the classes of the task. The second are a set of POP prompts that are learned continually, i.e. across all tasks, and encourage the model to discriminate {\it globally\/}, between all classes of all tasks. All learned prompts produce a corresponding set of tokens at the model output, which are fused to create task-based representations $f_t$ and a global representation $f$. A suitable set of losses is then used to optimize these representations for local and global performance. The representations are finally concatenated to create the feature space of the CIL classifier. 

Extensive experiments are presented to demonstrate the effectiveness of the POP model. Compared to classic trained-from-scratch methods, the foundation model based POP relies on a much more general substrate on which to learn each task. Hence, it achieves significantly better performance. Furthermore, because POP can use as few as a single prompt per task, low-shot continual learning becomes possible. With only 20 training samples per class, POP reaches results equivalent or superior to classic methods trained on the entire dataset. When compared to recent foundational model based CL methods, POP is shown to achieve better performance with fewer training examples. 

To sum up, the contributions of the paper are three fold:
\begin{enumerate}
    \item We propose POP, which leverages a pretrained foundation model to solve the CL problem, by learning a combination of local (task-specific) prompt sets and a global POP set. 
    \item We conduct extensive experiments to show that, POP reaches SOTA performance in the class incremental learning problem.
    \item We further show that POP ] unlocks the possibility of low-shot CL, or even memory buffer only CL, outperforming previous methods even in these challenging setups.
\end{enumerate}

\section{Related Work}
\vspace{-10pt}
{\bf Continual Learning:}
Continual Learning (CL) aims to learn sequences of tasks. The  goal is to learn new tasks without catastrophic forgetting of previously learned ones~\cite{survey1,survey2}. It can be further subdivided into the sub-problems of task incremental learning (TIL), domain incremental learning (DIL) and class incremental learning (CIL)~\cite{three}. CL has been studied in various areas of machine learning, including image classification~\cite{icarl, podnet, der}, video classification~\cite{pivot}, semantic segmentation~\cite{plop}, language analysis~\cite{progprompt}, vision-and-language tasks~\cite{climb}, etc. Nevertheless, the most popular CL testbed is image classification.

Gradient descend based neural networks are very vulnerable to catastrophic forgetting. To prevent this, the CL literature has proposed two main solutions: \textit{Distillation Based Methods} and \textit{Network Expansion Based Methods}. Distillation  methods distill knowledge learned by a model from old tasks (referred as the old model) into its learning of a new task (upon which it is referred to as the new model), so that the new model will not forget the old tasks. For this, iCaRL~\cite{icarl} first proposed to minimize the KL-Divergence between the logits of old and new models. Following this idea, LUCIR~\cite{lucir} and PODNet~\cite{podnet} proposed to conduct distillation over the feature vectors instead of logits. EWC~\cite{ewc}, GEM~\cite{gem} and NSCL~\cite{nscl}, on the other hand, suggested to constrain the gradient and parameters of the model, so as to remain close to those of the  old model. While distillation based methods can transfer knowledge to some extent, this transfer is eventually overwhelmed by the information learned from new tasks. Network expansion, or parameter isolation, methods were proposed to guarantee that old task knowledge is not forgotten. For this, RPSNet~\cite{rps} and DER~\cite{der} proposed to train a sub model per task,  which is an expert for this task and frozen in subsequent learning steps. The output features from all sub models are then integrated through averaging or concatenation. A limitation of these methods is the fast growth of model size, since a new network is added per task. To mitigate this, ProgressiveNet~\cite{PNN} and DNE~\cite{dne} proposed to add cross-task connections from old to new sub-models, to increase the efficiency of the forward knowledge transfer, allowing the addition of much smaller networks per task. Nevertheless, while network expansion methods can perform very well, their increasing model size and inference time, on the number of tasks can be problematic. Several strategies, such as network pruning~\cite{der}, or parameter sharing~\cite{pretrain, dne}, have been proposed to enable a better accuracy-complexity trade-off. 

{\bf Foundation Models:}
The recent introduction of foundation models promises a significant shift in various domains of machine learning. These are large-scale models trained with extremely large amounts of data~\cite{foundation_survey}, usually in a self-supervised manner. Popular such models include CLIP~\cite{clip}, GPT~\cite{gpt}, StableDiffusion~\cite{ldm}, SAM~\cite{sam}, etc. In many cases, these models can be applied to downstream tasks with none (zero-shot) or little (few-shot) training data, with very good performance. Since their training data covers a very wide range of categories and domains, their feature representations contain rich and diverse semantic information, enabling applications such as open-set classification. This makes foundation models naturally suited to CL by network expansion, where a foundation model can be used to initialize the sub-models learned per CL task. This makes the feature representations of the sub-models naturally related to each other, enabling the mitigation of the complexity of network expansion by parameter sharing, namely via prompt tuning methods.

{\bf Prompt Tuning:}
The rich feature space of a foundation model is a good starting point for learning new tasks. However, the adaptation to a  new task with limited data can destroy the quality of this feature representation. In general, naively fine-tuning the foundation model on downstream tasks is a poor solution. Prompt tuning~\cite{prefix_tuning, prompt_tuning, p_tuning, vpt} has been proposed to avoid this problem. This leverages the fact that most foundation models are transformers~\cite{transformer}, whose output feature representation can be adapted by tuning the attention mechanism alone. Prompt tuning methods fix the parameters of the foundation model, simply adding a small set of trainable tokens to accomplish this goal. Since these prompts have small learning capacity, the approach enables adaptation without catastrophic forgetting of the original foundation model representation. Recently, a few prompt tuning methods have been proposed. L2P~\cite{l2p} maintains a prompt pool, using the input image to find the prompts most related to the current task. DualPrompt~\cite{dual} first learns a group of general prompts and then adds prompts selected from a prompt pool. While prompt tuning enables these models to learn better task-specific representations, CIL also requires feature that encode cross-task information. The proposed POP approach further leverages prompt-tuning and  the transformer attention mechanism for this purpose. 


\section{Prompt of Prompts}
\vspace{-10pt}
\subsection{Continual Learning}
In CL, a  model $\mathcal{M}_\theta$, parameterized by $\theta$, learns a sequence of tasks $\{\mathcal{T}_1, \mathcal{T}_2, \dots, \mathcal{T}_T\}$. At step $k$, learning is based on dataset $\mathcal{D}_k$ of task $\mathcal{T}_k$. Previous datasets $\mathcal{D}_i, i<k$ are either unavailable or sub-sampled into a small memory buffer $\mathcal{B}$. The goal is to learn without forgetting, i.e. the model must perform well on both the new ($\mathcal{T}_k$) and old  ($\mathcal{T}_i, i<k$) tasks. Class Incremental Learning (CIL) is one of the most popular CL settings. It defines task $\mathcal{T}_k$ as a classification problem over a class label set $\mathcal{Y}_k$. While a classifier is learned from dataset $\mathcal{D}_k=\{(x_i,y_i)|(x_i,y_i)\sim P_k(x, y), y\in\mathcal{Y}_k\}_{i=1}^{N_k}$ of $N_k$ samples from distribution $P_k(X, Y)$, plus maybe a memory buffer $\cal B$ of data from previous tasks, the model is expected to produce a posterior distribution over all classes of all learned tasks, i.e. $\mathcal{M}^k_\theta(x)=P(y|x; \theta), y \in \cup_{i=1}^k \mathcal{Y}_i$.

Given the scarcity of data from previous tasks, it is challenging to maintain a good feature representation for those tasks, for models trained by gradient descent. In fact, the forgetting of the previously learned feature representation is the core problem of CL. One of the most popular and effective solutions is network expansion. At step $t+1$, the models learned in previous tasks, $\mathcal{M}_{\theta_i}, i \leq t$, are frozen and a new model, $\mathcal{M}_{\theta_{t+1}}$, is learned from $\mathcal{D}_{t+1}$. The input $x$ is then processed by all models, to produce a sequence of feature vectors
\begin{equation}
    f_i = \mathcal{M}_{\theta_i}(x),\quad i=1,2,\dots,t+1
\end{equation}
where $f_i$ is the feature representation of $x$ under model $\mathcal{M}_{\theta_i}$. Since each model $\mathcal{M}_{\theta_i}(x)$ is trained on a large dataset $\mathcal{D}_i$ of the $i^{th}$ task and then fixed, inputs $x$ from the classes of task $i$ can still be well represented by feature vector $f_i$, after step $t+1$. Hence by combining all $f_i, i \leq {t+1}$, the CL model can obtain all \textit{necessary} information to represent any input. This combination is usually implemented with the \textit{feature concatenation} operation $\oplus$
\begin{equation}
\label{equ:feat_cat}
    f = f_1\oplus f_2\oplus\dots\oplus f_{t+1}. \label{eq:concat}
\end{equation}

It is worth noting that, while $f$ contains all the \textit{necessary} information to represent any input, it is not naturally able to \textit{distinguish} the inputs from different tasks. Specifically, while $f_i$ can represent an input $x$ from task $i$, its behavior is unpredictable for inputs from tasks $j>i$, as model $\mathcal{M}_{\theta_i}$ is not trained with sample from task $j$. Some further enhancements, e.g. class balanced tuning and additional loss functions, are typically used to improve the performance of the concatenated feature vector $f$~\cite{der, dne}. Nevertheless, a strong drawback of this strategy is the need to both preserve the models learned for all tasks and process the input $x$ with all these models. The overall model size and inference time thus grow linearly with the number of tasks. Since the CL model should ideally be able to learn an unbounded sequence of tasks, the network expansion strategy will eventually exhaust all available memory and computation. Some methods, such as parameter sharing and network pruning, have been proposed to mitigate this problem~\cite{piggyback, pretrain, rps}.

\subsection{Foundation Models and Prompt Tuning}
In this work, we consider a different direction, based on foundation models, such as CLIP~\cite{clip}. A transformer-based foundation model $\mathcal{F}$ consists of a sequence of $L$ transformer~\cite{transformer} blocks $\mathcal{F}=\mathcal{F}_L\circ\mathcal{F}_{L-1}\circ\dots\circ\mathcal{F}_1$. The input $x$ is first subject to an embedding $\mathcal{E}$ into a set of $n$ tokens,
\begin{equation}
    [s_1, s_2, \dots, s_n] = \mathcal{E}(x),
\end{equation}
which are combined with a positional encoding $e_i$ and a class token $s_0$, and fed to ${\cal F}$, to generate a set of $n$ output tokens
\begin{equation}
    [r_0, r_1, r_2, \dots, r_n] = \mathcal{F}([s_0+e_0, s_1+e_1, s_2+e_2, \dots, s_n+e_n]).
\end{equation}
This can be written as 
\begin{equation}
    [r_0, R] = \mathcal{F}([s_0+e_0, S]) \label{eq:transformer}
\end{equation}
where $R=[r_1, r_2, \dots, r_n], S = [s_1+e_1, s_2+e_2, \dots, s_n+e_n]$. The class token output $f(x)=r_0$ is considered as the feature representation of $x$ for the solution of downstream tasks. 

Foundation models are large and trained on massive datasets, making the feature representation $f(x)$ quite generic and robust, and thus applicable to many downstream tasks. This allows the application of foundation models as zero-shot classifiers, capable of solving new classification tasks without any training. While the zero-shot classifier is not as good as a classifier trained on a large dataset of the new task, its performance can be surprisingly good~\cite{clip}. This has spurred interest on techniques for the adaptation of foundations models to new tasks. Classically, the adaptation of a model trained on dataset A to a new task B is achieved by fine-tuning the model on a dataset of B. This, however, is undesirable for foundation models, where A is massive. Fine-tuning the model on B destroys the representation $f(x)$ learned on A, sacrificing all the generality and robustness of this representation. 

{\it Prompt tuning\/} is an adaptation technique with better trade-off between the size of the adaptation dataset B and the performance of the adapted model. It consists of freezing the foundation model $\mathcal{F}$, and introducing a set of trainable tokens, $P=[p_1, p_2, \dots, p_m]$, known as {\it prompts\/}, at the transformer input, i.e. replacing (\ref{eq:transformer}) by
\begin{equation}
    [r_0, R, R_P] = \mathcal{F}([s_0+e_0, S, P]) \label{eq:spt}
\end{equation}
where $R_P=[r_{p_1}, r_{p_2}, \dots, r_{p_m}]$ and $r_{p_i}$ is the output corresponding to $p_i$. This is referred as the \textit{shallow prompt tuning} (SPT) strategy~\cite{vpt}. It is also possible to add prompts to each transformer block
\begin{equation}
    [s^i_0, S^i, S^i_P] = \mathcal{F}_i([s^{i-1}_0, S^{i-1}, P^{i-1}]), \quad \quad  s^0_0=s_0+e_0, S^0=S,
\end{equation}
and  define the model outputs as
\begin{equation}
    r_0=s^L_0, R=S^L, R_P=S^L_P,
\end{equation}
where $S^i=[s^i_1, s^i_2, \dots, s^i_n], S^i_P=[s^i_{p_1}, s^i_{p_2}, \dots, s^i_{p_m}]$, $s^i_j$ is the $j$-th output of the $i$-th block (and the the $j$-th input of $(i+1)$-th block). $P^{i}=[p^i_1, p^i_2, \dots, p^i_m]$ is an array of trainable prompts introduced at the input of $i$-th block. This is denoted as \textit{deep prompt tuning} (DPT)~\cite{vpt}. Note that, although the transformer block generates outputs $S^i_P$ corresponding to $P^i$, these are not used by subsequent transformer blocks. The transformer block is implemented with an attention block and a multi-layer perceptron (MLP) block~\cite{vit}. Since the attention block computes the correlations between all input tokens, the addition of prompts can change the semantics of the existing tokens. Hence, by learning prompts on dataset B, it is possible to adapt the model learned on A to new the task B. However, because the learning capacity is small (the model itself is frozen, only prompts are learned) this adaptation does not destroy the feature representation initially learned by the model on A.

\subsection{Foundational Feature Concatenation}
\label{sec:ffc}
The simplicity of adapting a transformer-based foundation model to new tasks by prompt-tuning suggests a natural solution to the CL problem. This is to learn a set of prompts $P_t$ per task $t$, using the corresponding dataset $\mathcal{D}_t$ (plus a memory buffer $\mathcal{B}$), and then somehow combine these prompts to adapt the transformer to {\it all\/} tasks. Taking SPT as an example, $P_t$ is learned by replacing (\ref{eq:spt}) by
\begin{equation}
\label{equ:foundation_cat1}
    [r_0, R, R_{P_t}] = \mathcal{F}([s_0+e_0, S, P_t]). \label{eq:promptcat}
\end{equation}
$P_t$ is then learned with
\begin{equation}
    P_t = \arg\min_{P_t} \frac{1}{N_t}\sum_{(x_i, y_i)\in\mathcal{D}_t\cup\mathcal{B}}\mathcal{L}(R_{P_t}(x_i), y_i)
\end{equation}
where $\cal L$ is the CIL loss function, and the feature representation of task $t$ is extracted from $R_{P_t}$. We use a simple average,
\begin{equation}
\label{equ:foundation_cat2}
    f_t = Mean(R_{P_t}). \label{eq:mean}
\end{equation}
but other operators are possible. One possibility to combine the individual task representations $f_t$, is to use (\ref{eq:concat}). This is the foundational model analogue of the feature concatenation strategy.  Since it leverages a frozen foundation model, which is shared across all tasks, only relying on the prompt sets $P_t$ to adapt it to each task, this strategy could in principle  have a better accuracy-complexity trade-off than traditional network expansion methods, which relearn a full model per task. 

\subsection{Prompt of Prompts}
\label{sec:pop}

While powerful, the concatenation strategy above ignores an important requirement of CL, which is to transfer knowledge from old to new tasks. Since representation $R_{P_t}$ is learned mostly from samples of task $t$, there could be substantial redundancy between representations of different tasks. Since prompts have, by design, limited capacity, this could hamper the learning of new concepts.  To force the learning of prompt set $P_t$ to account for the previously learned $P_i, i < t$, we propose to learn prompt sets sequentially. At step $t$, prompt sets $P_i, i < t$ are frozen and fed to the model together with learnable prompt set 
$R_t$, i.e (\ref{eq:promptcat}) is replaced by
\begin{equation}
    [r_0, R, R_{P_1}, R_{P_2}, \dots, R_{P_t}] = \mathcal{F}([s_0+e_0, S, P_1, P_2, \dots, P_t]). \label{eq:promptseq}
\end{equation}
This encourages  prompt set $P_t$, and the corresponding representation $R_{P_t}$ to contain information from task $t$ not already available in the previous task representations. In this way, the overall model adaptation capacity grows over time, since $R_{P_t}$ can reuse all features learned by previous tasks, through the transformer attention. Hence, over time, the model can learn more complex inter-task reasoning, specializing the adaptation according to what is really novel about the new tasks. However, because the previous features remain grounded by the previous tasks (frozen prompts) the  increase in capacity is limited to the set $P_t$, as is usual for prompt-tuning. This prevents the catastrophic forgetting of the representation initially available in the foundation model $\mathcal{F}$. 

It is, however, less clear how to combine the representations learned by different tasks. Note that while the prompts $P_i, i<t$ are frozen, the representations $R_{P_i}, i<t$ are not, since these also account for the input $x$ and prompt set $P_t$ through the attention layers of the model. The concatenation strategy would require the replacement of (\ref{eq:mean}) by
\begin{equation}
    f_t = Mean(R_{P_t}, R_{P_{t-1}}, \ldots, R_{P_1}), \label{eq:ftall}
\end{equation}
which would again mix the representation with those of previous tasks. To avoid this, we further propose to integrate the information across tasks using the foundation model itself. We propose to learn an extra group prompts over the prompts of all tasks, which is referred to as the set of {\it Prompt of Prompts \/} and denoted as $POP$. This is implemented by replacing (\ref{eq:promptseq}) with
\begin{equation}
    [r_0, R, R_{POP}, R_{P_1}, R_{P_2}, \dots, R_{P_t}] = \mathcal{F}([s_0+e_0, S, POP, P_1, P_2, \dots, P_t]).
\end{equation}
While each prompt set $P_t$ is learned uniquely from each task $t$, and then frozen, the $POP$ set is {\it continually learned} across all tasks, so as to integrate information across them. A cross-task feature representation is then obtained with
\begin{equation}
\label{equ:cross_feature}
    f_c = Mean(R_{POP}) \label{eq:fc}
\end{equation}
and combined with the task specific representations $f_1, f_2, \dots, f_t$, which can also be useful, by concatenation to obtain the final feature representation
\begin{equation}
\label{equ:pop_feature}
    f = f_1\oplus \dots\oplus f_t\oplus f_c \label{eq:fcat}
\end{equation}

\subsection{Training objectives}
Similarly to the popular DER~\cite{der} approach to CIL, we consider the combination of a CIL loss that encourages $R_{POP}$ to discriminate between all classes of all tasks, and an auxiliary loss that encourages each representation $R_{P_t}$ to be discriminative for the classes in   task $t$.  At step $t$, given an input $(x,y)$ from dataset $\mathcal{D}_k$, the auxiliary loss is defined as
\begin{equation}
    \mathcal{L}_{aux} = CE(\phi_{aux}(f_t(x)), \hat{y}) \quad \quad f_t(x) = Mean(R_{P_t}(x))
\end{equation}
where $CE$ is the cross-entropy loss, $\phi_{aux}$ is a linear classifier, and
\begin{equation}
    \hat{y}=\left\{\begin{array}{cc}
        0, & k < t \\
        y+1. & k=t.
    \end{array}\right.
\end{equation}
The resulting prompt set $P_t$ encourages the representation $f_t(x)$ to map classes that do not belong to task $t$ into a single (none of the above) category, while discriminating between the classes of task $t$. The CIL loss then combines task and class identity losses
\begin{equation}
    \mathcal{L}_{task} = CE(\phi_{task}(f(x)), k) \quad \quad 
    \mathcal{L}_{class} = CE(\phi_{task}(f(x)), y)
\end{equation}
where $f$ is the feature representation of~(\ref{equ:pop_feature}), and $\phi_{task}, \phi_{class}$ are linear classifiers. The resulting POP set encourages the representation $f(x)$ to discriminate between tasks and between all classes of all tasks. The final objective is the weighted average of the three losses above
\begin{equation}
    \mathcal{L} = \mathcal{L}_{class}+\lambda_{task}\mathcal{L}_{task}+\lambda_{aux}\mathcal{L}_{aux}
\end{equation}
where $\lambda_{task}$ and $\lambda_{aux}$ are hyperparameters. 

Figure~\ref{fig:model} summarizes the POP approach to CIL. And embedding $\cal E$ maps image patches into tokens, which are complemented by learned prompt sets $P_1, \ldots, P_t$ and POP. At step $t$ only $P_t$ and $POP$ are updated. These token are fed to the foundational model, which produces corresponding output prompt sets $R_{P_t}$ and $R_{POP}$. These are fused into feature representations $f_t$, using (\ref{eq:ftall}), and $f$, using (\ref{eq:fc})-(\ref{eq:fcat}). Finally, the representations $f_t$ trained for discrimination of the classes of task $t$ and the representation $f$ for discrimination of all classes of all tasks.

\section{Experiments}
\vspace{-10pt}
In this section, we evaluate the POP model on CIL benchmarks and compare it to existing methods.
\vspace{-5pt}
\subsection{Experiment Setup}
\vspace{-5pt}
\textbf{Benchmarks.} Similar to L2P~\cite{l2p} and DualPrompt~\cite{dual}, we evaluate CIL methods on the \textit{Split CIFAR-100} and \textit{Split ImageNet-R} benchmarks\footnote{All datasets used in the paper were solely downloaded and evaluated by UC San Diego}. Split CIFAR-100 divides CIFAR100~\cite{cifar} into 10 tasks, each  containing 10 classes. While Split CIFAR-100 is one of the most common CL testbeds, it may produce overly optimistic results for foundation model methods,  since these models are pretrained on massive datasets, which, with high probability, cover the CIFAR-100 categories and type of imagery. To mitigate the problem, we further evaluate on the Split ImageNet-R benchmark. Split ImageNet-R is built from ImageNet-R~\cite{imagenetr}, which contains images of ImageNet classes under domain shift (e.g. cartoon, graffiti, origami). Hence, ImageNet-R data is less typical of the imagery used to train foundation models, and thus better suited to evaluate the CL ability of foundation model based methods. ImageNet-R is split into 10 tasks, each containing 20 classes.

\textbf{Baselines.} The POP model is compared to several classic CIL methods,  ER~\cite{er}, BiC~\cite{bic}, GDumb~\cite{gdumb}, Dark~\cite{dark}, CO2L~\cite{co2l} and DER~\cite{der}, and two recently proposed prompt-based foundational model methods,  L2P~\cite{l2p} and DualPrompt~\cite{dual}. 

\textbf{Foundation models.}  While~\cite{l2p,dual} use a a Vit~\cite{vit} model pretrained on ImageNet-21K~\cite{imagenet}, we believe that this can provide a distorted view of the benefits of the foundation model approach to CIL. This is because ImageNet-21K is a dataset of images only, which contains most of the classes that appear in CIFAR-100 and ImageNet-R. Furthermore, the model is trained to discriminate between those classes. So, although there may be some domain mismatch, most of what is required for CIL is already available in the pre-trained model. We believe that CLIP is a better foundation model to evaluate CIL performance, for two reasons. First, it is not trained explicitly on a superset of the classes of CIFAR-100 and ImageNet-R. While it is trained on a very large dataset, which likely includes images from the ImageNet classes, those images are diffused into a much larger and more diverse pool of data. This is likely to make CLIP less ``ImageNet-overfitted". Second, and most importantly, the model is not trained for image classification, instead using a contrastive loss for image-text alignment. This further reduces the potential overfit to ImageNet-like classification tasks and provides a more generic feature space, which is a more realistic substrate for CIL. The surprising finding that it actually works reasonably well as a feature space for zero-shot classification is one of the main reasons why foundational are attracting sizeable attention in the  literature. For these reasons, we use a Vit model from CLIP~\cite{clip} as foundation model. 


\textbf{Implementation details.} Unless otherwise noted, results are presented for prompt tuning of a single prompt token per task and a single POP token. Hyperparameters are set to $\lambda_{task}=\lambda_{aux}=1$ and memory buffer size $|\mathcal{B}|$ to 1,000 or 5,000 images.

\textbf{Performance metrics.} Let $T$ be the number of tasks, and $A_t$ the classification accuracy over all classes learned up to and including task $t$. The \textit{Average Accuracy} $AA=\frac{1}{T}\sum_{i=1}^TA_i$ evaluates the model performance during the entire training process.

\begin{figure}\RawFloats
\centering
\begin{minipage}{.5\linewidth}
  \footnotesize
    \setlength{\tabcolsep}{2pt}
    \begin{tabular}{c|cccc} \toprule
    & & & \multicolumn{2}{c}{$AA(\uparrow)$ } \\
        Method & Pretraining & Buffer & CIFAR & ImgNetR\\ \hline
        \hline
        \multicolumn{5}{|c|}{Classic } \\
        \hline
        \hline
        ER      & N/A   & 1000 & 67.87 & 55.13\\
                &       & 5000 & 82.53 & 65.18\\ \hline
        BiC     & N/A   & 1000 & 66.11 & 52.14\\
                &       & 5000 & 81.42 & 64.63\\ \hline
        GDumb   & N/A   & 1000 & 67.14 & 38.32\\
                &       & 5000 & 81.67 & 65.90\\ \hline
        Dark    & N/A   & 1000 & 61.06 & 55.47\\
                &       & 5000 & 83.94 & 66.73\\ \hline
        CO2L    & N/A   & 1000 & 72.15 & 53.45\\
                &       & 5000 & 82.49 & 65.90\\ \hline
                
        DER     & N/A   & 2000 & 74.64 & \\ \hline
      \multicolumn{5}{|c|}{Foundational  } \\
                   \hline
        \hline
        L2P     & ImageNet-21K   & 0    & 83.86 & 61.57\\
                &                & 1000 & 84.21 &\\
                &                & 5000 & 86.31 &\\ \hline
        DualPrompt & ImageNet-21K& 0    & \textbf{86.51} & 68.13\\ \midrule\midrule
        POP     & CLIP           & 1000 & 81.97 & 74.22\\
                &                & 5000 & 85.75& \textbf{79.74}\\\bottomrule
    \end{tabular}
  \captionof{table}{$AA$ on Split CIFAR-100 and Split ImageNet-R for both classic and foundation model-based CL approaches.}
    \label{tab:results}
\end{minipage}
\hspace{.1in}
\begin{minipage}{.45\linewidth}
  \includegraphics[width=0.9\linewidth]{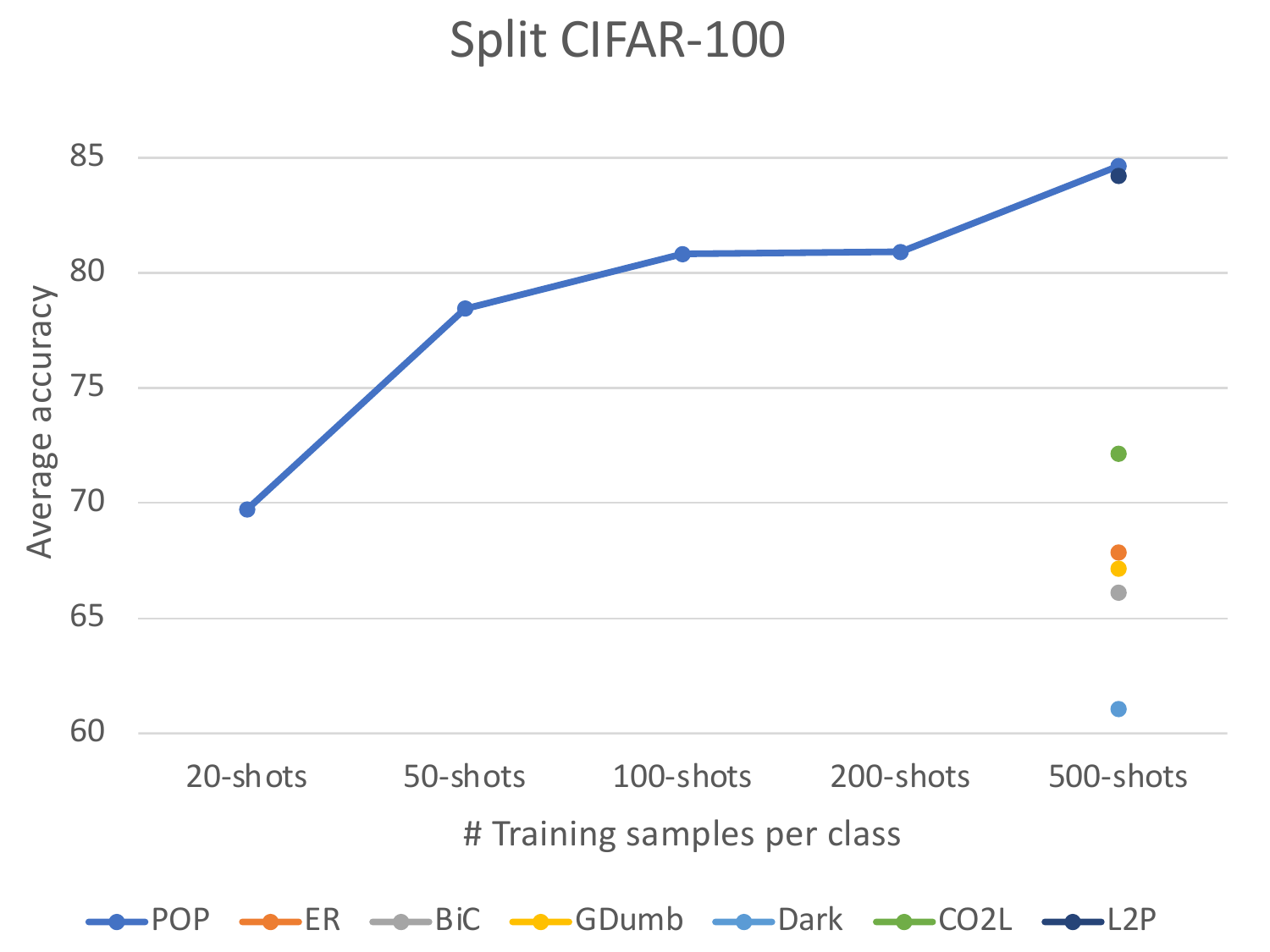}
  \includegraphics[width=0.9\linewidth]{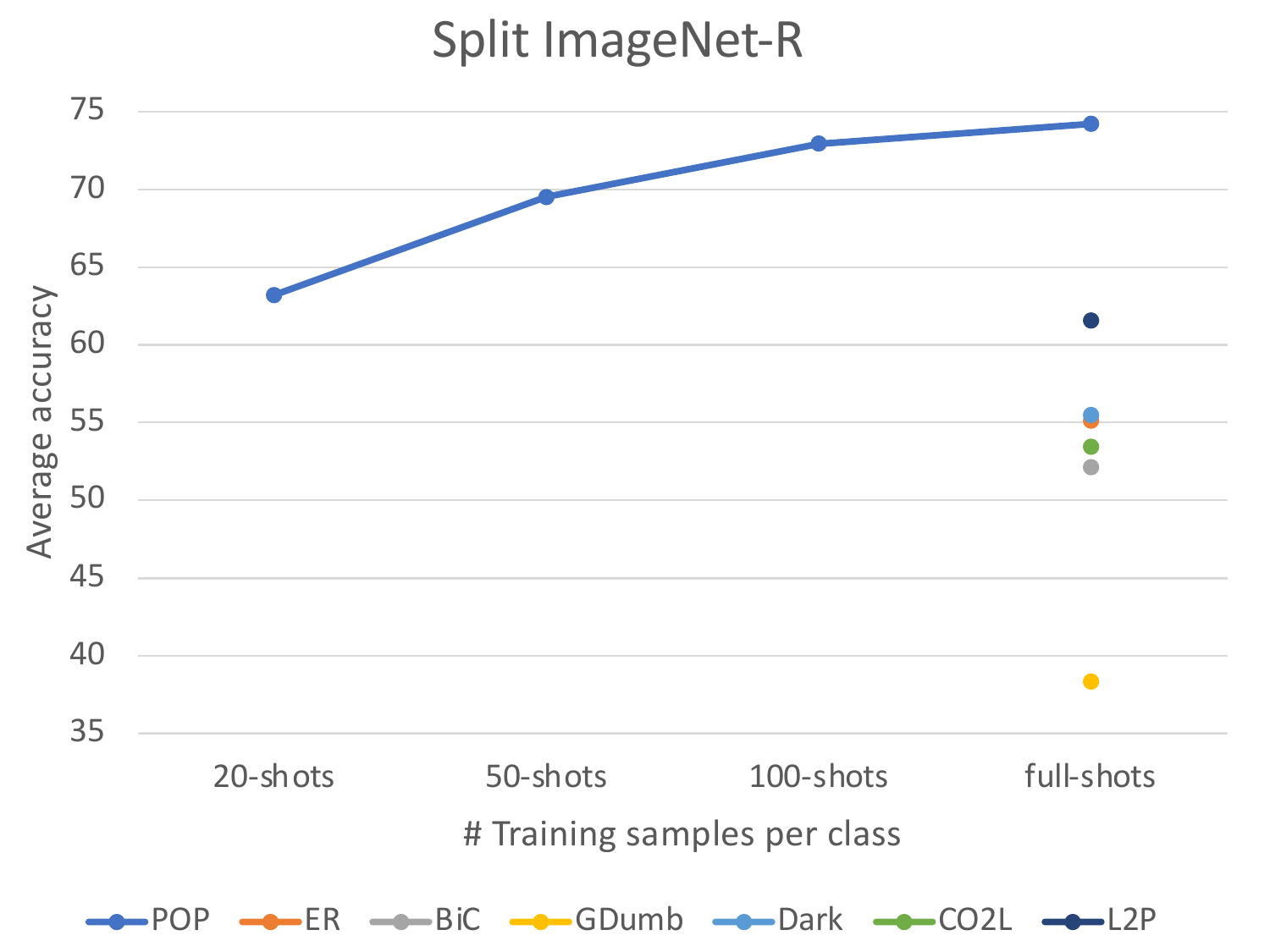}
  \caption{POP model performance in the $k$-shot setting, as a function of $k$.}
  \label{fig:fewshot_results}
\end{minipage}
\end{figure}

%
\vspace{-5pt}
\subsection{Results}
\vspace{-5pt}
We start by comparing POP to classic CIL methods (ER, BiC, GDumb, Dark, CO2L, DER). Table~\ref{tab:results} shows that POP is clearly superior. For buffer size 1,000, the SOTA classic method achieves 72.15\%/55.47\% AA on Split CIFAR-100/ImageNet-R. POP outperforms these rates by 9.82\%/18.75\%. While the gap reduces to 1.81\%/13.01\% for the larger buffer of size 5,000, it is still substantial. 

We next compare to L2P and DualPrompt which are prompt based methods that also use a foundation model. On Split CIFAR-100, POP underperforms these methods.  We argue that, because the categories of CIFAR-100 are contained in ImageNet-21K, the ViT trained on ImageNet-21K can naturally represent CIFAR-100 categories. This is supported  by the change of accuracy of the various methods with memory size.  For classic methods, increasing the memory size from 1,000 to 5,000 can improve performance by 10\%-20\%. For the CLIP-based POP, performance rises by almost $4\%$. However, for the ImageNet-21K-trained ViT of L2P this improvement is merely 2.1\%. 

A much different picture emerges for ImageNet-R, where L2P and DualPrompt have performance on par or only slightly superior to the classic approaches. For example, CO2L with buffer size 5,000 achieves $65.9\%$, which is superior to the results reported for L2P~\footnote{Although these are achieved with a buffer size of 0, which is the only setting it reports for this dataset.} and close to the $68.13\%$ of DualPrompt. On this dataset, the proposed implementation of POP, using the CLIP ViT model, achieves a significantly higher 74.22\% for buffer size 1,000 and an even higher 79.74\% for buffer size 5,000. This is almost $15\%$ higher than the classical SOTA. These results support the claim that the features of the CLIP ViT are more more generic and thus suitable for CIL, but also harder to adapt to CIFAR-100. To eliminate all doubts, we implemented POP using the ViT pretrained on ImageNet-21K and obtained accuracies of 84.64/89.00 with memory size 1,000/5,000 on Split CIFAR-100. These results are superior to those obtained by all methods, and significantly superior to those of POP with the CLIP ViT. This further confirms the ``ImageNet overfitting'' problem of the ImageNet-21K model.



\subsection{Low-shot Continual Learning}
\vspace{-5pt}

\begin{figure}
    \centering
    \includegraphics[width=0.35\linewidth]{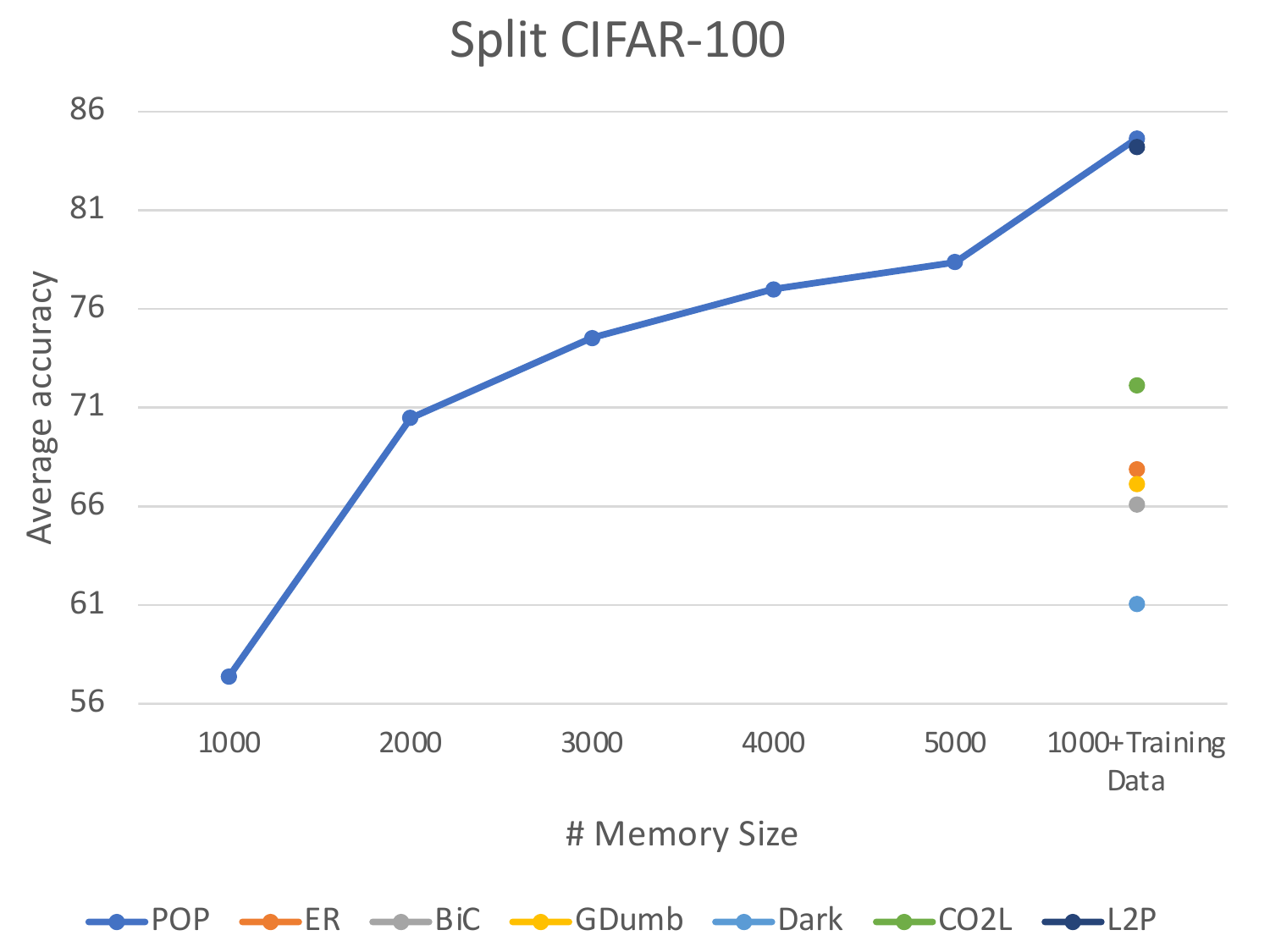}
    \includegraphics[width=0.35\linewidth]{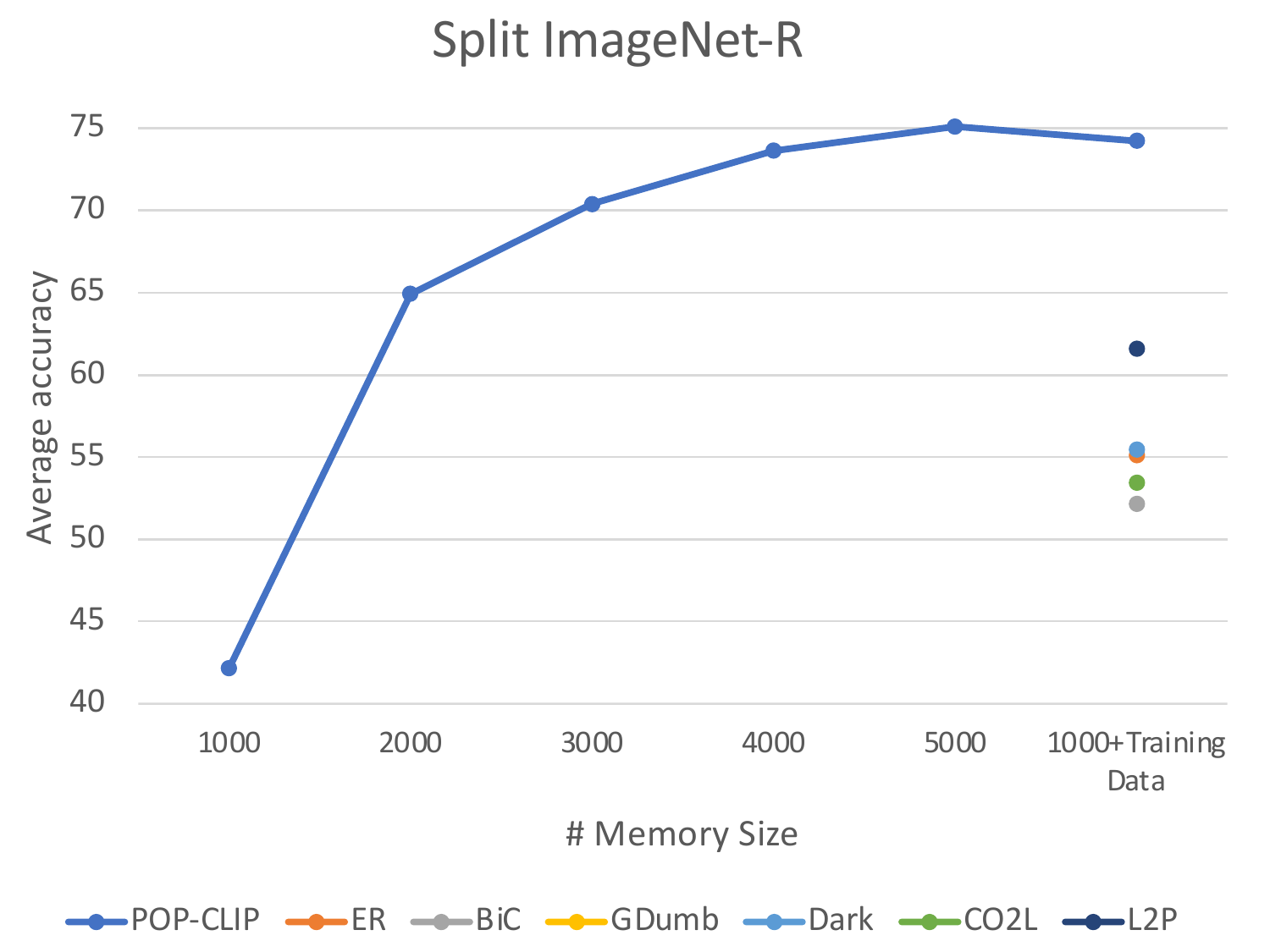}
    \caption{Performances of the proposed POP model that only utilize the memory buffer.}
    \label{fig:mem_only_results}
\end{figure}

We next consider the problem of $k$-shot CIL, where only $k$ training samples are available per class in the dataset $\mathcal{D}_t$ used to train task $t$. Figure~\ref{fig:fewshot_results} illustrates the performance of  POP for the CLIP-ViT foundational model. The buffer size is 1,000 (10 images per class for CIFAR-100, 5 for ImageNet-R) and $k$ increases from $k=20$ to the available number of training samples. On Split CIFAR-100, with only 50 samples per task, POP already outperforms all classic methods, which are trained with 500 samples per class. On Split ImageNet-R, POP outperforms both classic and prompt-based methods with only 20 samples per class, while remaining methods use an average of 150 per class.

Traditionally, the main difficulty of CIL is the scarcity of data from previous tasks. Although classical methods maintain a memory buffer, its small size prevented the possibility of solving the CIL problem by simply learning from this buffer. This is different for foundation model approaches, such as POP, which can effectively learn in the few-shot setting. Hence, it should be possible to learn from the memory buffer alone. Figure~\ref{fig:mem_only_results} reports to the setting where the POP model uses the limited  examples available in the buffer uniquely to learn a joint classifier over all categories. Results are presented as a function of buffer size $|\mathcal{B}|$, which ranges from 1000 to 5000. This accounts for 2\% to 10\% of the CIFAR-100 dataset or 3.3\% to 16.7\% of the ImageNet-R dataset. On Split CIFAR-100, POP outperform all classic methods with a buffer of 3000 samples (30 per class), on Split ImageNet-R with a buffer of 2000 samples (10 per class).
\vspace{-5pt}
\subsection{Ablation Study}
\vspace{-5pt}

\begin{figure}
\begin{floatrow}
\ffigbox{%
  \includegraphics[width=0.7\linewidth]{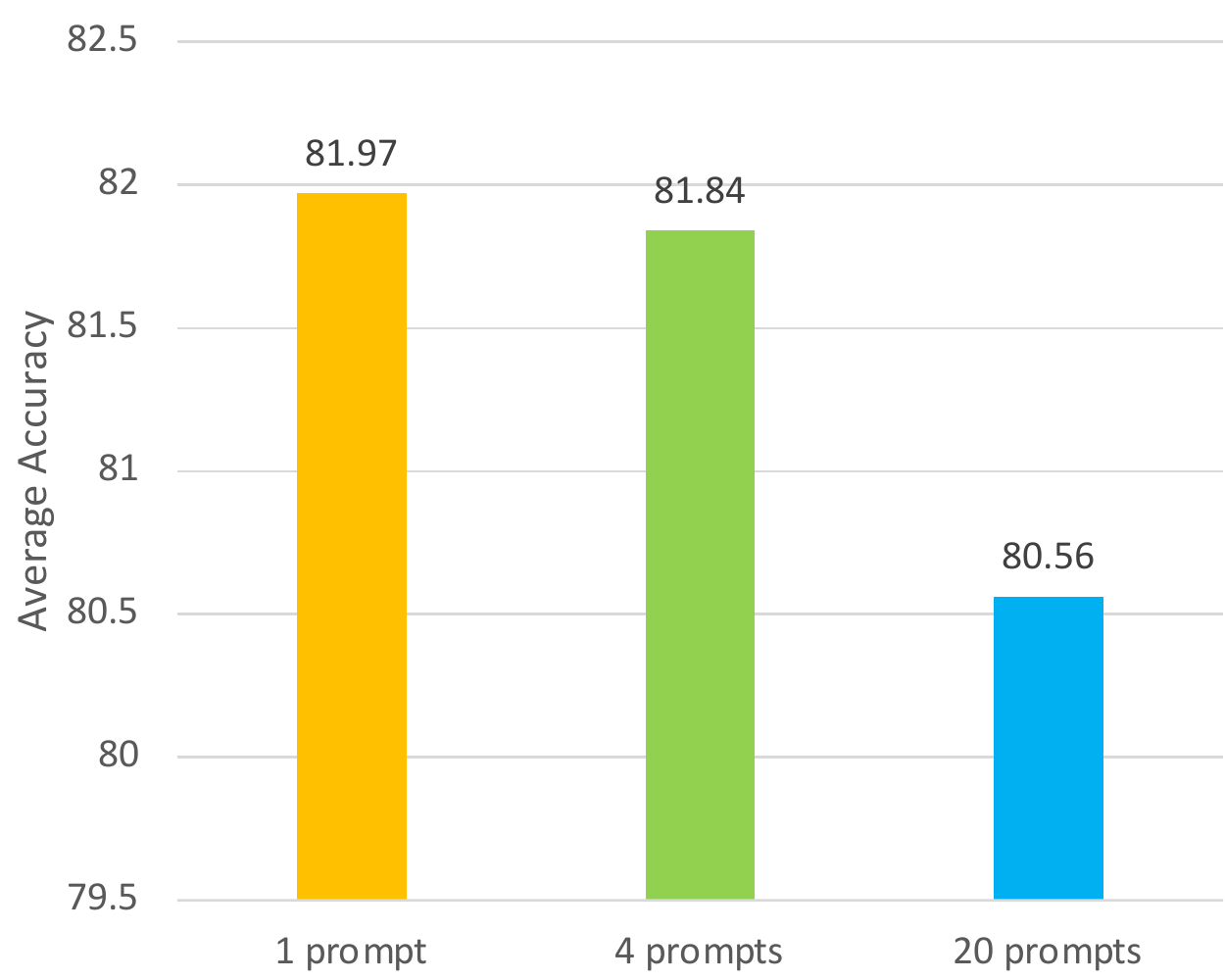}%
}{%
  \caption{How the number of prompts affects the performances of POP}%
  \label{fig:prompt_num}
}
\capbtabbox{%
  \begin{tabular}{c|c} \toprule
        Feature Fusing Method & $AA(\uparrow)$\\ \hline
        FF cat & 79.68\\
        Mean of all & 80.81\\
        Max pooling & 80.92\\
        POP token only & 80.13\\
        POP (Mean and cat) & \textbf{81.97}\\ \bottomrule
    \end{tabular}
    
}{%
  \caption{Different feature fusing methods for POP}%
  \label{tab:fuse}
}
\end{floatrow}
\end{figure}

\textbf{Number of prompts per task.} We evaluate how the number of prompts learned per task affects CIL performance. We use the POP model with ViT from CLIP and conduct the experiments on Split CIFAR-100 with memory size 1000. Figure~\ref{fig:prompt_num} shows that learning one prompt per task yields the best performance, with a slight performance degradation when more prompts are used. In any case, the performance does not vary significantly between 1 and 4 prompts.

\textbf{Feature fusing methods} The proposed POP model averages the POP output features in (\ref{equ:cross_feature}) and concatenates them in (\ref{equ:pop_feature}). This denoted as the mean and cat method. Table~\ref{sec:ffc} lists other possibilities for feature fusion. FF cat refers to the foundation feature concatenation combining (\ref{equ:feat_cat}), (\ref{equ:foundation_cat1}), and (\ref{equ:foundation_cat2}). It is also possible to directly compute the mean or max pooling of all features, as in (\ref{eq:ftall}), or only use the representation derived from POP tokens, i.e. (\ref{eq:fc}) 
instead of (\ref{eq:fcat}). More detailed descriptions are included in the supplementary material. Among all these feature fusing methods, mean and cat has the highest performance.

\vspace{-10pt}
\section{Conclusion}
\vspace{-10pt}
In this work, we propose to leverage the strong representation ability of foundation models to solve the CL problem. Leveraging the effectiveness of prompt tuning, we propose the Prompt Of Prompts (POP) model, which adapts the foundation model to different tasks with a group of task-specific prompts, and integrates knowledge across all tasks with a group of global POP prompts. The POP model is shown to outperform classical CL methods by a large margin. Compared to other foundation model based methods, POP uses a more generic foundational backbone, which enables less overfiting to ImageNet-style imagery and better CL performance in the presence of domain shift. Moreover, we have shown that POP can successfully perform low-shot CL. With as few as 50 samples per class (memory buffer size 3000), it outperforms previous methods trained on the entire dataset.

{\bf Limitations:} While foundational models, like POP, are being shown highly effective for CL, it is currently difficult to isolate image datasets that have not been used in their training. This is likely to result in optimistic estimates of CL performance. On the other hand, as foundational models are trained on ever larger datasets, they may eventually cover the entire natural image distribution. The practical significance of CIL results then depends on considerations such as the one that we have made above, namely the use of models trained for generic representation learning, like CLIP. More research is needed to fully understand these questions.

{\bf Social Impact:} POP-based CL does not pose any specific social concerns, but inherits all the concerns usually associated with deep learning systems, such as vulnerability to attacks, dataset biases, etc., which the CL setting may amplify.

{\small
\bibliographystyle{unsrtnat}
\bibliography{ref}

\begin{thebibliography}{45}
\providecommand{\natexlab}[1]{#1}
\providecommand{\url}[1]{\texttt{#1}}
\expandafter\ifx\csname urlstyle\endcsname\relax
  \providecommand{\doi}[1]{doi: #1}\else
  \providecommand{\doi}{doi: \begingroup \urlstyle{rm}\Url}\fi

\bibitem[Rebuffi et~al.(2017{\natexlab{a}})Rebuffi, Kolesnikov, Sperl, and
  Lampert]{icarl}
Sylvestre-Alvise Rebuffi, Alexander Kolesnikov, Georg Sperl, and Christoph~H
  Lampert.
\newblock icarl: Incremental classifier and representation learning.
\newblock In \emph{Proceedings of the IEEE conference on Computer Vision and
  Pattern Recognition}, pages 2001--2010, 2017{\natexlab{a}}.

\bibitem[Douillard et~al.(2020)Douillard, Cord, Ollion, Robert, and
  Valle]{podnet}
Arthur Douillard, Matthieu Cord, Charles Ollion, Thomas Robert, and Eduardo
  Valle.
\newblock Podnet: Pooled outputs distillation for small-tasks incremental
  learning.
\newblock In \emph{Computer Vision--ECCV 2020: 16th European Conference,
  Glasgow, UK, August 23--28, 2020, Proceedings, Part XX 16}, pages 86--102.
  Springer, 2020.

\bibitem[Buzzega et~al.(2020)Buzzega, Boschini, Porrello, Abati, and
  Calderara]{dark}
Pietro Buzzega, Matteo Boschini, Angelo Porrello, Davide Abati, and Simone
  Calderara.
\newblock Dark experience for general continual learning: a strong, simple
  baseline.
\newblock \emph{Advances in neural information processing systems},
  33:\penalty0 15920--15930, 2020.

\bibitem[Hou et~al.(2019)Hou, Pan, Loy, Wang, and Lin]{lucir}
Saihui Hou, Xinyu Pan, Chen~Change Loy, Zilei Wang, and Dahua Lin.
\newblock Learning a unified classifier incrementally via rebalancing.
\newblock In \emph{Proceedings of the IEEE/CVF conference on Computer Vision
  and Pattern Recognition}, pages 831--839, 2019.

\bibitem[Kirkpatrick et~al.(2017)Kirkpatrick, Pascanu, Rabinowitz, Veness,
  Desjardins, Rusu, Milan, Quan, Ramalho, Grabska-Barwinska, et~al.]{ewc}
James Kirkpatrick, Razvan Pascanu, Neil Rabinowitz, Joel Veness, Guillaume
  Desjardins, Andrei~A Rusu, Kieran Milan, John Quan, Tiago Ramalho, Agnieszka
  Grabska-Barwinska, et~al.
\newblock Overcoming catastrophic forgetting in neural networks.
\newblock \emph{Proceedings of the national academy of sciences}, 114\penalty0
  (13):\penalty0 3521--3526, 2017.

\bibitem[Wang et~al.(2021)Wang, Li, Sun, and Xu]{nscl}
Shipeng Wang, Xiaorong Li, Jian Sun, and Zongben Xu.
\newblock Training networks in null space of feature covariance for continual
  learning.
\newblock In \emph{Proceedings of the IEEE/CVF conference on Computer Vision
  and Pattern Recognition}, pages 184--193, 2021.

\bibitem[Zhou et~al.(2023{\natexlab{a}})Zhou, Li, Li, Yu, Liu, Wang, Zhang, Ji,
  Yan, He, et~al.]{foundation_survey}
Ce~Zhou, Qian Li, Chen Li, Jun Yu, Yixin Liu, Guangjing Wang, Kai Zhang, Cheng
  Ji, Qiben Yan, Lifang He, et~al.
\newblock A comprehensive survey on pretrained foundation models: A history
  from bert to chatgpt.
\newblock \emph{arXiv preprint arXiv:2302.09419}, 2023{\natexlab{a}}.

\bibitem[Brown et~al.(2020)Brown, Mann, Ryder, Subbiah, Kaplan, Dhariwal,
  Neelakantan, Shyam, Sastry, Askell, et~al.]{gpt}
Tom Brown, Benjamin Mann, Nick Ryder, Melanie Subbiah, Jared~D Kaplan, Prafulla
  Dhariwal, Arvind Neelakantan, Pranav Shyam, Girish Sastry, Amanda Askell,
  et~al.
\newblock Language models are few-shot learners.
\newblock \emph{Advances in neural information processing systems},
  33:\penalty0 1877--1901, 2020.

\bibitem[Rombach et~al.(2022)Rombach, Blattmann, Lorenz, Esser, and Ommer]{ldm}
Robin Rombach, Andreas Blattmann, Dominik Lorenz, Patrick Esser, and Bj{\"o}rn
  Ommer.
\newblock High-resolution image synthesis with latent diffusion models.
\newblock In \emph{Proceedings of the IEEE/CVF Conference on Computer Vision
  and Pattern Recognition}, pages 10684--10695, 2022.

\bibitem[Kirillov et~al.(2023)Kirillov, Mintun, Ravi, Mao, Rolland, Gustafson,
  Xiao, Whitehead, Berg, Lo, et~al.]{sam}
Alexander Kirillov, Eric Mintun, Nikhila Ravi, Hanzi Mao, Chloe Rolland, Laura
  Gustafson, Tete Xiao, Spencer Whitehead, Alexander~C Berg, Wan-Yen Lo, et~al.
\newblock Segment anything.
\newblock \emph{arXiv preprint arXiv:2304.02643}, 2023.

\bibitem[Vaswani et~al.(2017)Vaswani, Shazeer, Parmar, Uszkoreit, Jones, Gomez,
  Kaiser, and Polosukhin]{transformer}
Ashish Vaswani, Noam Shazeer, Niki Parmar, Jakob Uszkoreit, Llion Jones,
  Aidan~N Gomez, {\L}ukasz Kaiser, and Illia Polosukhin.
\newblock Attention is all you need.
\newblock \emph{Advances in neural information processing systems}, 30, 2017.

\bibitem[Lester et~al.(2021)Lester, Al-Rfou, and Constant]{prompt_tuning}
Brian Lester, Rami Al-Rfou, and Noah Constant.
\newblock The power of scale for parameter-efficient prompt tuning.
\newblock \emph{arXiv preprint arXiv:2104.08691}, 2021.

\bibitem[Zhou et~al.(2023{\natexlab{b}})Zhou, Wang, Qi, Ye, Zhan, and
  Liu]{survey1}
Da-Wei Zhou, Qi-Wei Wang, Zhi-Hong Qi, Han-Jia Ye, De-Chuan Zhan, and Ziwei
  Liu.
\newblock Deep class-incremental learning: A survey.
\newblock \emph{arXiv preprint arXiv:2302.03648}, 2023{\natexlab{b}}.

\bibitem[Qu et~al.(2021)Qu, Rahmani, Xu, Williams, and Liu]{survey2}
Haoxuan Qu, Hossein Rahmani, Li~Xu, Bryan Williams, and Jun Liu.
\newblock Recent advances of continual learning in computer vision: An
  overview.
\newblock \emph{arXiv preprint arXiv:2109.11369}, 2021.

\bibitem[Van~de Ven and Tolias(2019)]{three}
Gido~M Van~de Ven and Andreas~S Tolias.
\newblock Three scenarios for continual learning.
\newblock \emph{arXiv preprint arXiv:1904.07734}, 2019.

\bibitem[Yan et~al.(2021)Yan, Xie, and He]{der}
Shipeng Yan, Jiangwei Xie, and Xuming He.
\newblock Der: Dynamically expandable representation for class incremental
  learning.
\newblock In \emph{Proceedings of the IEEE/CVF Conference on Computer Vision
  and Pattern Recognition}, pages 3014--3023, 2021.

\bibitem[Villa et~al.(2022)Villa, Alc{\'a}zar, Alfarra, Alhamoud, Hurtado,
  Heilbron, Soto, and Ghanem]{pivot}
Andr{\'e}s Villa, Juan~Le{\'o}n Alc{\'a}zar, Motasem Alfarra, Kumail Alhamoud,
  Julio Hurtado, Fabian~Caba Heilbron, Alvaro Soto, and Bernard Ghanem.
\newblock Pivot: Prompting for video continual learning.
\newblock \emph{arXiv preprint arXiv:2212.04842}, 2022.

\bibitem[Douillard et~al.(2021)Douillard, Chen, Dapogny, and Cord]{plop}
Arthur Douillard, Yifu Chen, Arnaud Dapogny, and Matthieu Cord.
\newblock Plop: Learning without forgetting for continual semantic
  segmentation.
\newblock In \emph{Proceedings of the IEEE/CVF Conference on Computer Vision
  and Pattern Recognition}, pages 4040--4050, 2021.

\bibitem[Razdaibiedina et~al.(2023)Razdaibiedina, Mao, Hou, Khabsa, Lewis, and
  Almahairi]{progprompt}
Anastasia Razdaibiedina, Yuning Mao, Rui Hou, Madian Khabsa, Mike Lewis, and
  Amjad Almahairi.
\newblock Progressive prompts: Continual learning for language models.
\newblock \emph{arXiv preprint arXiv:2301.12314}, 2023.

\bibitem[Srinivasan et~al.(2022)Srinivasan, Chang, Pinto~Alva, Chochlakis,
  Rostami, and Thomason]{climb}
Tejas Srinivasan, Ting-Yun Chang, Leticia Pinto~Alva, Georgios Chochlakis,
  Mohammad Rostami, and Jesse Thomason.
\newblock Climb: A continual learning benchmark for vision-and-language tasks.
\newblock \emph{Advances in Neural Information Processing Systems},
  35:\penalty0 29440--29453, 2022.

\bibitem[Lopez-Paz and Ranzato(2017)]{gem}
David Lopez-Paz and Marc'Aurelio Ranzato.
\newblock Gradient episodic memory for continual learning.
\newblock \emph{Advances in neural information processing systems}, 30, 2017.

\bibitem[Rajasegaran et~al.(2019)Rajasegaran, Hayat, Khan, Khan, and Shao]{rps}
Jathushan Rajasegaran, Munawar Hayat, Salman~H Khan, Fahad~Shahbaz Khan, and
  Ling Shao.
\newblock Random path selection for continual learning.
\newblock \emph{Advances in Neural Information Processing Systems}, 32, 2019.

\bibitem[Rusu et~al.(2016)Rusu, Rabinowitz, Desjardins, Soyer, Kirkpatrick,
  Kavukcuoglu, Pascanu, and Hadsell]{PNN}
Andrei~A Rusu, Neil~C Rabinowitz, Guillaume Desjardins, Hubert Soyer, James
  Kirkpatrick, Koray Kavukcuoglu, Razvan Pascanu, and Raia Hadsell.
\newblock Progressive neural networks.
\newblock \emph{arXiv preprint arXiv:1606.04671}, 2016.

\bibitem[Hu et~al.(2023)Hu, Li, Lyu, Gao, and Vasconcelos]{dne}
Zhiyuan Hu, Yunsheng Li, Jiancheng Lyu, Dashan Gao, and Nuno Vasconcelos.
\newblock Dense network expansion for class incremental learning.
\newblock \emph{arXiv preprint arXiv:2303.12696}, 2023.

\bibitem[Wu et~al.(2022)Wu, Swaminathan, Li, Ravichandran, Vasconcelos,
  Bhotika, and Soatto]{pretrain}
Tz-Ying Wu, Gurumurthy Swaminathan, Zhizhong Li, Avinash Ravichandran, Nuno
  Vasconcelos, Rahul Bhotika, and Stefano Soatto.
\newblock Class-incremental learning with strong pre-trained models.
\newblock In \emph{Proceedings of the IEEE/CVF Conference on Computer Vision
  and Pattern Recognition}, pages 9601--9610, 2022.

\bibitem[Radford et~al.(2021)Radford, Kim, Hallacy, Ramesh, Goh, Agarwal,
  Sastry, Askell, Mishkin, Clark, et~al.]{clip}
Alec Radford, Jong~Wook Kim, Chris Hallacy, Aditya Ramesh, Gabriel Goh,
  Sandhini Agarwal, Girish Sastry, Amanda Askell, Pamela Mishkin, Jack Clark,
  et~al.
\newblock Learning transferable visual models from natural language
  supervision.
\newblock In \emph{International conference on machine learning}, pages
  8748--8763. PMLR, 2021.

\bibitem[Li and Liang(2021)]{prefix_tuning}
Xiang~Lisa Li and Percy Liang.
\newblock Prefix-tuning: Optimizing continuous prompts for generation.
\newblock \emph{arXiv preprint arXiv:2101.00190}, 2021.

\bibitem[Liu et~al.(2021)Liu, Ji, Fu, Tam, Du, Yang, and Tang]{p_tuning}
Xiao Liu, Kaixuan Ji, Yicheng Fu, Weng~Lam Tam, Zhengxiao Du, Zhilin Yang, and
  Jie Tang.
\newblock P-tuning v2: Prompt tuning can be comparable to fine-tuning
  universally across scales and tasks.
\newblock \emph{arXiv preprint arXiv:2110.07602}, 2021.

\bibitem[Jia et~al.(2022)Jia, Tang, Chen, Cardie, Belongie, Hariharan, and
  Lim]{vpt}
Menglin Jia, Luming Tang, Bor-Chun Chen, Claire Cardie, Serge Belongie, Bharath
  Hariharan, and Ser-Nam Lim.
\newblock Visual prompt tuning.
\newblock In \emph{Computer Vision--ECCV 2022: 17th European Conference, Tel
  Aviv, Israel, October 23--27, 2022, Proceedings, Part XXXIII}, pages
  709--727. Springer, 2022.

\bibitem[Wang et~al.(2022{\natexlab{a}})Wang, Zhang, Lee, Zhang, Sun, Ren, Su,
  Perot, Dy, and Pfister]{l2p}
Zifeng Wang, Zizhao Zhang, Chen-Yu Lee, Han Zhang, Ruoxi Sun, Xiaoqi Ren,
  Guolong Su, Vincent Perot, Jennifer Dy, and Tomas Pfister.
\newblock Learning to prompt for continual learning.
\newblock In \emph{Proceedings of the IEEE/CVF Conference on Computer Vision
  and Pattern Recognition}, pages 139--149, 2022{\natexlab{a}}.

\bibitem[Wang et~al.(2022{\natexlab{b}})Wang, Zhang, Ebrahimi, Sun, Zhang, Lee,
  Ren, Su, Perot, Dy, et~al.]{dual}
Zifeng Wang, Zizhao Zhang, Sayna Ebrahimi, Ruoxi Sun, Han Zhang, Chen-Yu Lee,
  Xiaoqi Ren, Guolong Su, Vincent Perot, Jennifer Dy, et~al.
\newblock Dualprompt: Complementary prompting for rehearsal-free continual
  learning.
\newblock In \emph{Computer Vision--ECCV 2022: 17th European Conference, Tel
  Aviv, Israel, October 23--27, 2022, Proceedings, Part XXVI}, pages 631--648.
  Springer, 2022{\natexlab{b}}.

\bibitem[Mallya et~al.(2018)Mallya, Davis, and Lazebnik]{piggyback}
Arun Mallya, Dillon Davis, and Svetlana Lazebnik.
\newblock Piggyback: Adapting a single network to multiple tasks by learning to
  mask weights.
\newblock In \emph{Proceedings of the European Conference on Computer Vision
  (ECCV)}, pages 67--82, 2018.

\bibitem[Dosovitskiy et~al.(2020)Dosovitskiy, Beyer, Kolesnikov, Weissenborn,
  Zhai, Unterthiner, Dehghani, Minderer, Heigold, Gelly, et~al.]{vit}
Alexey Dosovitskiy, Lucas Beyer, Alexander Kolesnikov, Dirk Weissenborn,
  Xiaohua Zhai, Thomas Unterthiner, Mostafa Dehghani, Matthias Minderer, Georg
  Heigold, Sylvain Gelly, et~al.
\newblock An image is worth 16x16 words: Transformers for image recognition at
  scale.
\newblock \emph{arXiv preprint arXiv:2010.11929}, 2020.

\bibitem[Krizhevsky et~al.(2009)Krizhevsky, Hinton, et~al.]{cifar}
Alex Krizhevsky, Geoffrey Hinton, et~al.
\newblock Learning multiple layers of features from tiny images.
\newblock 2009.

\bibitem[Hendrycks et~al.(2021)Hendrycks, Basart, Mu, Kadavath, Wang, Dorundo,
  Desai, Zhu, Parajuli, Guo, et~al.]{imagenetr}
Dan Hendrycks, Steven Basart, Norman Mu, Saurav Kadavath, Frank Wang, Evan
  Dorundo, Rahul Desai, Tyler Zhu, Samyak Parajuli, Mike Guo, et~al.
\newblock The many faces of robustness: A critical analysis of
  out-of-distribution generalization.
\newblock In \emph{Proceedings of the IEEE/CVF International Conference on
  Computer Vision}, pages 8340--8349, 2021.

\bibitem[Hayes et~al.(2019)Hayes, Cahill, and Kanan]{er}
Tyler~L Hayes, Nathan~D Cahill, and Christopher Kanan.
\newblock Memory efficient experience replay for streaming learning.
\newblock In \emph{2019 International Conference on Robotics and Automation
  (ICRA)}, pages 9769--9776. IEEE, 2019.

\bibitem[Wu et~al.(2019)Wu, Chen, Wang, Ye, Liu, Guo, and Fu]{bic}
Yue Wu, Yinpeng Chen, Lijuan Wang, Yuancheng Ye, Zicheng Liu, Yandong Guo, and
  Yun Fu.
\newblock Large scale incremental learning.
\newblock In \emph{Proceedings of the IEEE/CVF Conference on Computer Vision
  and Pattern Recognition}, pages 374--382, 2019.

\bibitem[Prabhu et~al.(2020)Prabhu, Torr, and Dokania]{gdumb}
Ameya Prabhu, Philip~HS Torr, and Puneet~K Dokania.
\newblock Gdumb: A simple approach that questions our progress in continual
  learning.
\newblock In \emph{Computer Vision--ECCV 2020: 16th European Conference,
  Glasgow, UK, August 23--28, 2020, Proceedings, Part II 16}, pages 524--540.
  Springer, 2020.

\bibitem[Cha et~al.(2021)Cha, Lee, and Shin]{co2l}
Hyuntak Cha, Jaeho Lee, and Jinwoo Shin.
\newblock Co2l: Contrastive continual learning.
\newblock In \emph{Proceedings of the IEEE/CVF International conference on
  computer vision}, pages 9516--9525, 2021.

\bibitem[Deng et~al.(2009)Deng, Dong, Socher, Li, Li, and Fei-Fei]{imagenet}
Jia Deng, Wei Dong, Richard Socher, Li-Jia Li, Kai Li, and Li~Fei-Fei.
\newblock Imagenet: A large-scale hierarchical image database.
\newblock In \emph{2009 IEEE conference on computer vision and pattern
  recognition}, pages 248--255. Ieee, 2009.

\bibitem[Netzer et~al.(2011)Netzer, Wang, Coates, Bissacco, Wu, and Ng]{svhn}
Yuval Netzer, Tao Wang, Adam Coates, Alessandro Bissacco, Bo~Wu, and Andrew~Y
  Ng.
\newblock Reading digits in natural images with unsupervised feature learning.
\newblock 2011.

\bibitem[Stallkamp et~al.(2012)Stallkamp, Schlipsing, Salmen, and Igel]{gtsrb}
Johannes Stallkamp, Marc Schlipsing, Jan Salmen, and Christian Igel.
\newblock Man vs. computer: Benchmarking machine learning algorithms for
  traffic sign recognition.
\newblock \emph{Neural networks}, 32:\penalty0 323--332, 2012.

\bibitem[Maji et~al.(2013)Maji, Rahtu, Kannala, Blaschko, and
  Vedaldi]{aircraft}
Subhransu Maji, Esa Rahtu, Juho Kannala, Matthew Blaschko, and Andrea Vedaldi.
\newblock Fine-grained visual classification of aircraft.
\newblock \emph{arXiv preprint arXiv:1306.5151}, 2013.

\bibitem[Soomro et~al.(2012)Soomro, Zamir, and Shah]{ucf101}
Khurram Soomro, Amir~Roshan Zamir, and Mubarak Shah.
\newblock Ucf101: A dataset of 101 human actions classes from videos in the
  wild.
\newblock \emph{arXiv preprint arXiv:1212.0402}, 2012.

\bibitem[Rebuffi et~al.(2017{\natexlab{b}})Rebuffi, Bilen, and
  Vedaldi]{decathlon}
Sylvestre-Alvise Rebuffi, Hakan Bilen, and Andrea Vedaldi.
\newblock Learning multiple visual domains with residual adapters.
\newblock \emph{Advances in neural information processing systems}, 30,
  2017{\natexlab{b}}.

\end{thebibliography}
}

\newpage

\noindent\textbf{\Large Appendix}
\appendix

\renewcommand{\thefigure}{A}

\section{Implementation Details}
In this section, we provide more implementation details of the POP model.

\subsection{Training setup of POP model}
To train the POP model, we use the Adam optimizer with a initial learning rate 5e-4, weight decay 1e-6 and batch size 64. The model is first trained for 170 epochs with the dataset of current task $\mathcal{D}_t$ and memory buffer $\mathcal{B}$. The learning rate is dropped by a factor of 10 at epoch 60, 100 and 120. After that, we use the class balanced tuning trick of~\cite{der} to enhance model performance. Specifically, we downsample $\mathcal{D}_t$ so that the categories in $\mathcal{D}_t$ and the categories in $\mathcal{B}$ have the same number of samples. The model is trained for 20 epochs in the class balanced tuning stage. The foundation model we use is a ViT-base~\cite{vit} model with patch size 16.

\subsection{Details of low-shot continual learning}
In continual learning, a sequence of tasks $\{\mathcal{T}_1, \mathcal{T}_2, \dots, \mathcal{T}_T\}$ and their corresponding training datasets $\mathcal{D}_1, \mathcal{D}_2, \dots \mathcal{D}_T$ is sequentially available. In low-shot continual learning, only a small subset of $\mathcal{D}'_t\subseteq \mathcal{D}_t (|\mathcal{D}'_t| \ll |\mathcal{D}_t|)$ is available for task $t$. In the literature, $\mathcal{D}'_t$ is commonly generated by randomly sub-sampling from $\mathcal{D}_t$. Low-shot learning is much more challenging than standard continual learning. Not only the samples of previous tasks, but also those of the current task, become low-shot. Without the support of a strong pretrained model it is usually impossible to solve the low-shot continual learning problem effectively.

\subsection{Learning with the buffer only}
Joint training, i.e. learning the model from the union of the datasets of all the tasks ($\mathcal{D}_1\cup\mathcal{D}_2\cup\dots\cup\mathcal{D}_T$) is an upper bound for continual learning performance. In the standard continual setting, where only a small buffer $\cal B$ of data is available from the previous tasks, joint learning is not possible. However, for low-shot continual learning, this becomes possible if the number of shots is low enough. For example, in the 5-shot setting, only 5 images are available per class of the new task. If the buffer contains 5 images from each of the classes in previous tasks, there is no point in using a ``continual learning" algorithm. The images of the new classes are simply added to buffer $\cal B$ and the model is updated by relearning all tasks jointly, using the data in $\cal B$. This is denoted as ``learning with the buffer only." By enabling low-shot learning, the use of foundation models and prompt tuning makes this type of learning feasible.

\subsection{Details of feature fusion methods}
In Table 4 of the main paper, we compare several feature fusion methods to the Mean-and-cat method in POP. We provide mathematical descriptions of these methods in the following.

\textbf{FF-cat.} The foundation feature concatenation method first learns a set of prompts per task, using (9) and (11) of the main paper. These task specific features are concatenated with (2) to generate the final feature vector $f$.

\textbf{Mean-of-all.} We first extract the features $R_{P_1}, R_{P_2}, \dots, R_{P_T}$  corresponding to the task prompts and the features $R_{POP}$ corresponding to the POP prompt, using (14) of the main paper. The final feature is computed as
\begin{equation}
    f = Mean(R_{P_1}, R_{P_2}, \dots, R_{P_T}, R_{POP})
\end{equation}

\textbf{Max-pooling.} Task specific features $f_1, f_2, \dots, f_T$ and cross task feature $f_c$ are first computed with (13) and (15). We then  take the maximum value of each position of the feature vectors.
The final feature vector is
\begin{equation}
    f[i] = max(f_1[i], f_2[i], \dots, f_T[i], f_c[i])
\end{equation}
where $f[i]$ is the $i$-th entry of $f$

\textbf{POP-token-only.} This method only use the output corresponding to the POP tokens to generate the final feature vector, i.e.
\begin{equation}
    f = f_c
\end{equation}

\section{Additional Experiments}

To further validate the effectiveness of the POP model, we provide comparisons to the prompt based L2P~\cite{l2p} model for additional datasets. We select the SVHN~\cite{svhn}, GTSRB~\cite{gtsrb}, Aircraft~\cite{aircraft} and UCF101~\cite{ucf101} datasets from the Visual Domain Decathlon challenge~\cite{decathlon}. Note that, in the Visual Domain Decathlon, the Aircraft and UCF101 tasks are already low-shot, Aircraft has an average of 33 samples per class and UCF101 an average of 76 samples per class. For GTSRB, Aircraft and UCF101, we split the categories into 10 tasks, for SVHN into 5 tasks. The memory buffer size is set to 1,000 for both POP and L2P. Since the L2P paper does not provide results on these datasets, we conducted the experiments ourselves, using the official implementation of L2P.

\begin{table}[htbp]
    \centering
    \begin{tabular}{c|cc|cccc} \toprule
    & & & \multicolumn{4}{c}{$AA(\uparrow)$ } \\
        Method & Pretraining & Buffer & SVHN & GTSRB & Aircraft & UCF101\\ \hline
        L2P     & ImageNet-21K   & 1000 & 69.65 & 84.79 & 22.14 & 30.57\\
        POP     & CLIP           & 1000 & \textbf{74.26} & \textbf{94.82} & \textbf{38.52} & \textbf{50.88}\\\bottomrule
    \end{tabular}
  \captionof{table}{$AA$ on SVHN, GTSRB, Aircraft and UCF101 for POP and L2P.}
    \label{tab:results}
\end{table}

Tabel~\ref{tab:results} summarizes the results from this experiment. POP outperforms L2P on all datasets. On the medium sized datasets, the gains are of 4.61\% on SVHN and 10.03\% on GTSRB. On the more challenging low-shot datasets, the gains are larger: POP outperforms L2P by 16.38\% on Aircraft and 20.31\% on UCF101. This further verifies our conclusion in the main paper: the features of the POP model are more generic and more amenable to low-shot learning.

\end{document}